\documentclass[11pt]{article}
\usepackage{amssymb,amsmath,bm,graphicx}
\usepackage{mathrsfs}
\usepackage{constants}
\usepackage{hyperref}
\usepackage{enumitem}
\usepackage{color}

\usepackage{float}
\restylefloat{table}
\usepackage{placeins}

\usepackage{amssymb}
\usepackage{graphicx}

\usepackage{subcaption}
\usepackage{caption}
\usepackage{csvsimple}
\usepackage[T2A]{fontenc}
\usepackage[utf8]{inputenc}
\usepackage{babel}
\usepackage{dblfloatfix}
\usepackage{hyperref}
\usepackage{amsmath}
\usepackage{lipsum}
\usepackage{amsfonts}
\usepackage{caption}

\usepackage{newtxmath,booktabs,sectsty}
\paragraphfont{\mdseries\itshape}
\usepackage{tabularx,ragged2e}
\newcolumntype{L}[1]{>{\RaggedRight\hsize=#1\hsize}X}
\newcolumntype{C}[1]{>{\Centering\hsize=#1\hsize\hspace{0pt}}X}
\newcommand\mycell[1]{\smash{%
  \begin{tabular}[t]{@{}>{\RaggedRight}p{\hsize}@{}} #1 \end{tabular}}}

\date{}
\begin{document}
\newcommand{\bea}{\begin{eqnarray}}
\newcommand{\ena}{\end{eqnarray}}
\newcommand{\beas}{\begin{eqnarray*}}
\newcommand{\enas}{\end{eqnarray*}}
\newcommand{\beq}{\begin{equation}}
\newcommand{\enq}{\end{equation}}
\def\qed{\hfill \mbox{\rule{0.5em}{0.5em}}}
\newcommand{\bbox}{\hfill $\Box$}
\newcommand{\ignore}[1]{}
\newcommand{\ignorex}[1]{#1}
\newcommand{\wtilde}[1]{\widetilde{#1}}
\newcommand{\qmq}[1]{\quad\mbox{#1}\quad}
\newcommand{\qm}[1]{\quad\mbox{#1}}
\newcommand{\nn}{\nonumber}
\newcommand{\Bvert}{\left\vert\vphantom{\frac{1}{1}}\right.}
\newcommand{\To}{\rightarrow}
\newcommand{\E}{\mathbb{E}}
\newcommand{\Var}{\mathrm{Var}}
\newcommand{\Cov}{\mathrm{Cov}}
\newcommand{\Corr}{\mathrm{Corr}}
\newcommand{\dist}{\mathrm{dist}}
\newcommand{\diam}{\mathrm{diam}}
\makeatletter
\newsavebox\myboxA
\newsavebox\myboxB
\newlength\mylenA
\newcommand*\xoverline[2][0.70]{%
    \sbox{\myboxA}{$\m@th#2$}%
    \setbox\myboxB\null
    \ht\myboxB=\ht\myboxA%
    \dp\myboxB=\dp\myboxA%
    \wd\myboxB=#1\wd\myboxA
    \sbox\myboxB{$\m@th\overline{\copy\myboxB}$}
    \setlength\mylenA{\the\wd\myboxA}
    \addtolength\mylenA{-\the\wd\myboxB}%
    \ifdim\wd\myboxB<\wd\myboxA%
       \rlap{\hskip 0.5\mylenA\usebox\myboxB}{\usebox\myboxA}%
    \else
        \hskip -0.5\mylenA\rlap{\usebox\myboxA}{\hskip 0.5\mylenA\usebox\myboxB}%
    \fi}
\makeatother

\newtheorem{theorem}{Theorem}[section]
\newtheorem{corollary}[theorem]{Corollary}
\newtheorem{conjecture}[theorem]{Conjecture}
\newtheorem{proposition}[theorem]{Proposition}
\newtheorem{lemma}[theorem]{Lemma}
\newtheorem{definition}[theorem]{Definition}
\newtheorem{example}[theorem]{Example}
\newtheorem{remark}[theorem]{Remark}
\newtheorem{case}{Case}[section]
\newtheorem{condition}{Condition}[section]
\newcommand{\proof}{\noindent {\it Proof:} }

\title{{\bf\Large Clustering performance analysis using a new correlation-based cluster validity index}}
\author{Nathakhun Wiroonsri \thanks{Email: nathakhun.wir@kmutt.ac.th }  \\ Mathematics and Statistics with Applications Research Group (MaSA) \\ Department of Mathematics, King Mongkut's University of Technology Thonburi}

\footnotetext{AMS 2010 subject classifications: Primary 62H30\ignore{Cluster Analysis} Secondary 68T10\ignore{Pattern recognition}.}

\maketitle

\begin{abstract}
There are various cluster validity indices used for evaluating clustering results. One of the main objectives of using these indices is to seek the optimal unknown number of clusters. Some indices work well for clusters with different densities, sizes, and shapes. Yet, one shared weakness of those validity indices is that they often provide only one optimal number of clusters. That number is unknown in real-world problems, and there might be more than one possible option. We develop a new cluster validity index based on a correlation between an actual distance between a pair of data points and a centroid distance of clusters that the two points occupy. Our proposed index constantly yields several local peaks and overcomes the previously stated weakness. Several experiments in different scenarios, including UCI real-world data sets, have been conducted to compare the proposed validity index with several well-known ones.  An R package related to this new index called NCvalid is available at \url{https://github.com/nwiroonsri/NCvalid}.

\end{abstract}

\textbf{Keyword}: clustering algorithm, cluster validity measure, correlation, data partitions, marketing, pattern recognition



\section{Introduction} \label{sec:introduction}

Cluster analysis is one of the most popular unsupervised tools in statistical and machine learning (see \cite{JTF09} and \cite{JWHT13} for an overview of the method.) Researchers apply it to solve problems in various fields, such as medicine, social science, image processing, and biology. Currently, in the world of big data, it plays a very significant role in marketing. One of the most known techniques is to categorize customers to determine an effective business strategy for each group, where the number of groups is unknown (see \cite{CAK08}, \cite{KV13} and references therein.) There is no true number of classes in this area, and there is more than one potential option. These facts motivated us to write this paper.  

There are two main goals of developing a new theory in cluster analysis.  The first one is to develop a new cluster algorithm, such as k-means \cite{Ste56}, \cite{Mac67}, Hierarchical clustering \cite{Sib73}, \cite{Def77}, fuzzy c-means \cite{Dun73}, \cite{Bez81}, EM algorithm \cite{DLR77}, DBSCAN \cite{EKSX96}, and many more recent techniques, including deep learning based methods. The second goal is to develop a new cluster measurement to evaluate the cluster quality or find the optimal number of clusters. This field has attracted great attention for about a half-century. Several works developed indices for crisp clustering as detailed later in Section \ref{sec:background}, and many works developed ones for fuzzy clustering (see  \cite{Win82}, \cite{XB91}, \cite{PB95}, \cite{ZZSY09}, \cite{PKBH11}, \cite{PKBH13}, \cite{ZWL19}, \cite{KKS21}, and \cite{NZB22} for example.) In this work, we introduce a new cluster validity index for crisp clustering based on a correlation coefficient, such as Pearson, Kendall, and Spearman between the distances of each pair of points and each pair of centroids of the clusters that the two points occupy. The correlation introduced here can also check whether the selected clustering result has met a user's expectation.

Several known cluster validity indices handle the cases where each cluster has a different shape, size, and density. The developed validity indices completed their performance checks based on known class data sets to see whether or not they indicate the true number of classes as the optimum. Some of those data sets have multiple classes that are very close as if they are from the same class regardless of the background knowledge. It could be that other factors affect the classes excluded from the data set. The natural question is, “How can we be certain that those indices provide an appropriate number of clusters if the true class is unknown in reality?” That is, switching to another specific application, assuming that the data has a similar pattern as just mentioned, “How can we be certain that the same pattern in this new application should separate into the same number of clusters?”

A problem here is that most indices sometimes give a clear optimum only at a specific number of clusters. Using this information, we tend to select the number of clusters at the peak, which may cause problems. For instance, in marketing, where customer segmentation is the goal, the true number of clusters is unknown and never will be known. Assume that three groups and six groups both give good partitions, and an experienced marketing analyst decides on a final number of groups. If the indices give only one peak, we will lose this information. 

This problem motivates us to propose a new cluster validity index that always provides several local peaks with different heights so users can rank and choose the number of clusters appropriate for their applications. In some situations, breaking observations into too few or too many groups is not suitable, and thus a sub-optimal option should suffice. The proposed index uses a correlation coefficient between the distances of each pair of points and the corresponding centroids of the clusters that the two points occupy. This correlation was inspired by cophenetic correlation \cite{SR62}, which is used only for hierarchical clustering. Our correlation is quite similar to point biserial correlation \cite{Mil80} with a binary label (0/1 for same/different cluster) replaced by the actual distance between the centroids of clusters that the two points occupy. We then define some adjustments to the correlation to get our validity index. The works \cite{PKBH11} and \cite{PKBH13} also introduced correlation-based validity indices for fuzzy clustering in a much more complicated formation. Nevertheless, they are not intended to handle the problem raised in the previous paragraphs. Though our new indices are defined simply, they outperform the well-known ones in many aspects, especially for crisp clustering.

Next, we present two examples that correspond to the stated problem above.

\begin{example} \label{24ex}

\begin{figure}[h]
    \centering
    \begin{subfigure}[t]{0.47\textwidth} 
		    \centering
        \includegraphics[width=1.5in]{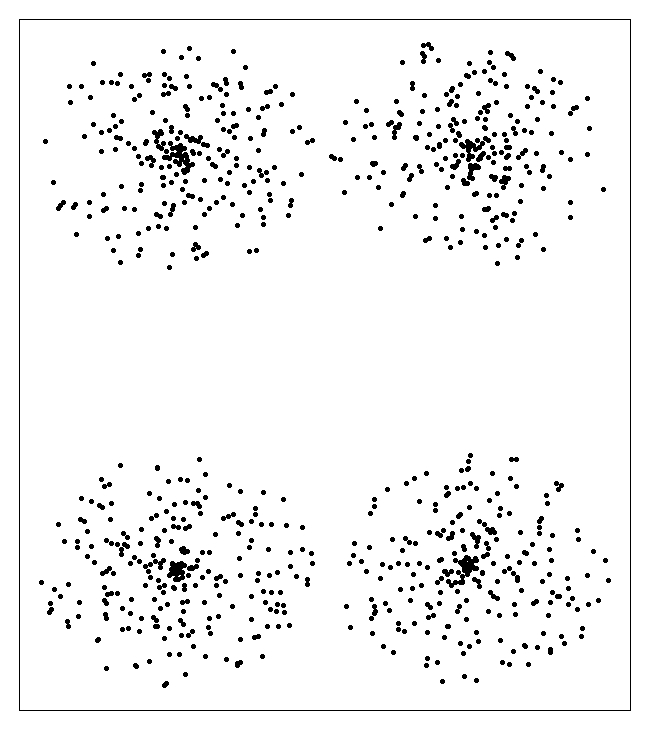}
        \caption{2 or 4 groups?}
        \label{2or4}
    \end{subfigure}
    \begin{subfigure}[t]{0.47\textwidth} 
		     \centering
         \includegraphics[width=1.8in]{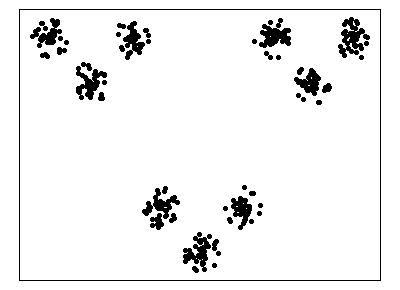}
         \caption{3 or 9 groups?}
         \label{3or9}   
    \end{subfigure}
\caption{Examples for our index's highlighted feature}
\label{fig:main}
\end{figure}

The data set adapted from \cite{CSL04} shown in Figure \ref{2or4} is generated from four uniform distributions with 250 points each. It is clear that the best partition is by dividing into either four or two groups, and four is preferable in this case as the two consecutive groups look different. In some situations, a user may wish to separate the data points into only two groups. We will see later that most well-known indices give the optimal number at four, but two is not even an option, while our index finds two local peaks at two and four. 
\end{example}

\begin{example} \label{39ex}
The data set shown in Figure \ref{3or9} is generated from nine uniform distributions with 50 data points each. Notably, the best partition is either three big groups or nine small groups. So, if we do not know the true labels, we might want to break them into three or nine groups, with six being a secondary choice. Several known validity indices give clear peaks only at either three or nine. We will see later that our proposed indices give peaks at three, six, and nine with different heights. 
\end{example}

To verify that our proposed validity index can handle the problem raised above well, we evaluate and compare them to ten well-known validity indices on several artificial and real-world benchmark data sets.  The ten validity indices are Calinski-Harabasz \cite{CH74}, Chou-Su-Lai(CS) \cite{CSL04}, Dunn’s index \cite{Dun73}, Davies-Bouldin’s index \cite{DB79}, Davies-Bouldin* index(DB*) \cite{KR05}, Generalized Dunn index (GD33) \cite{BP98}, point biserial correlation \cite{Mil80}, PBM index \cite{PBM04}, silhouette coefficient \cite{Rou87},\cite{KR90}, and STR index \cite{Sta17}.

The remainder of this paper is organized as follows. Section \ref{sec:background} provides the necessary background of the cluster validity indices. Then, we state our new cluster validity index in Section \ref{sec:main}. In Section \ref{sec:comp}, we compare computational costs of the selected validity indices. The experimental results are presented in Section \ref{sec:exp}. Finally, we devote Section \ref{sec:conclude} for discussion and conclusion. 

\section{Background of cluster validity indices}\label{sec:background}
In this section, we provide some background and definitions of validity indices for crisp clustering that we compare with our new validity index. For more cluster validity indices, an overview and two comparative experiments appear in \cite{DJ79}, \cite{MC85}, and \cite{AGMPP13}, respectively. 

In this work, we let $\{x_1,x_2,\ldots,x_n\}$ be a data set of size $n \in \mathbb{N}$  where $x_i = (x_{i1},x_{i,2},\ldots,x_{ip})$ with $p \in \mathbb{N}$ be the number of attributes. Let $\vec{d} = (d(x_i,x_j))_{i,j \in [n]}$ be a vector of length ${n \choose 2}$ containing distances of all pairs of points, and let $d(\cdot,\cdot)$ be a distance function (e.g., Euclidean) where $[n]=\{1,2,\ldots,n\}$. Also, we denote $k$ as the number of clusters unless otherwise stated and denote $C_i$ for $i=1,2,\ldots,k$ as the set of data points in the $i$\textsuperscript{th} cluster and $v_i=(v_{i1},v_{i2},\ldots,v_{ip})$ as the centroid of the $i$\textsuperscript{th} cluster. Also let $\bar{x}$ be the global centroid.  

Next, we provide the definitions of the ten selected cluster validity indices. 

\textbf{
\begin{flushleft}
Calinski-Harabasz (CH)\cite{CH74}
\end{flushleft}
}
The Calinski-Harabasz index is one of the most classic cluster validity indices defined as
\beas
\texttt{CH}(k) = \frac{n-k}{k-1}\frac{\sum_{i=1}^k|C_i|d(v_i,\bar{x})}{\sum_{i=1}^k\sum_{x_j\in C_i}d(x_j,v_i)}.
\enas
The numerator measures the distances between centroids of the clusters and the global centroid, and the denominator measures within-cluster distances to centroids. Clearly, the largest $\texttt{CH}(k)$ indicates a valid optimal partition. 

\begin{flushleft}
\textbf{Chou-Su-Lai measure (CS) \cite{CSL04}}
\end{flushleft}
The CS measure is defined as
\beas
\texttt{CS}(k) = \frac{\sum_{i=1}^k \left\{\frac{1}{|C_i|}\sum_{x_j \in C_i} \max_{x_l \in C_i} d(x_j,x_l)\right\}}{\sum_{i=1}^k \left\{\min_{j:j \ne i}d(v_i,v_j)\right\}}.
\enas
The numerator is the sum of the average maximum within-cluster distance of each point over all clusters, and the denominator is the sum of the minimum between-cluster separation of each cluster. The smallest $\texttt{CS}(k)$ indicates a valid optimal partition.

\begin{flushleft}
\textbf{Dunn's index (DI) \cite{Dun73}}
\end{flushleft}

Dunn's index is a classic and effective validity index defined as
\beas
\texttt{DI}(k) = \min_{i \ne j \in [k]}\left\{\frac{\min\left\{d(x_u,x_v)|x_u\in C_i,x_v \in C_j\right\}}{\max_{l \in [k]}\max\left\{d(x_u,x_v)|x_u,x_v \in C_l\right\}}\right\}.
\enas

The numerator represents the minimum between-cluster distance, and the denominator represents the maximum within-cluster distance. The largest $\texttt{DI}(k)$ indicates a valid optimal partition.

\begin{flushleft}
\textbf{Davies-Bouldin's index (DB) \cite{DB79}}
\end{flushleft}

To define Davies-Bouldin's index, we begin with the within-$i$\textsuperscript{th}-cluster scatter and the between-$i$\textsuperscript{th}-and-$j$\textsuperscript{th}-cluster distance. For $q,t \ge 1$ and $i\ne j \in [k]$, let
\beas
S_{i,q} = \left(\frac{1}{|C_i|}\sum_{x \in C_i}\left\|x-v_i\right\|_2^q\right)^{1/q} \text{ \ and \ } M_{ij,t} = \left(\sum_{s=1}^p|v_{is}-v_{js}|^t\right)^{1/t}.
\enas
Then, let
\beas
R_{i,qt} = \max_{j \in [k]\textbackslash \left\{i\right\}}\left\{\frac{S_{i,q}+S_{j,q}}{M_{ij,t}}\right\}.
\enas

The Davies-Bouldin's index is defined as
\beas
\texttt{DB}(k) = \frac{1}{k}\sum_{i=1}^kR_{i,qt}.
\enas
The smallest $\texttt{DB}(k)$ indicates a valid optimal partition.
 
\begin{flushleft}
\textbf{Davies-Bouldin* index (DB*) \cite{KR05}}
\end{flushleft}

The DB* index is defined similarly to DB with $R_{i,qt}$ replaced by
\beas
R^*_{i,qt} = \frac{\max_{j \in [k]\textbackslash \left\{i\right\}}\left\{S_{i,q}+S_{j,q}\right\}}{\min_{j \in [k]\textbackslash \left\{i\right\}}M_{ij,t}}.
\enas

This adjustment of Davies-Bouldin's index originated in \cite{KR05}. Similar to the original, the smallest $\texttt{DB*}(k)$ indicates a valid optimal partition.

\begin{flushleft}
\textbf{Generalized Dunn index 33 (GD33) \cite{BP98}}
\end{flushleft}
The generalization of the Dunn index originated in \cite{BP98}, where they proposed 18 variations. We select GD33 as it performs best in our tested scenarios and in the experiments in \cite{AGMPP13}. It is defined as
\beas
\texttt{GD33}(k) = \min_{i \ne j \in [k]}\left\{\frac{\frac{1}{|C_i||C_j|}\sum_{x_u \in C_i,x_v\in C_j}d(x_u,x_v)}{\max_{l \in [k]}\frac{2}{|C_l|}\sum_{x_u \in C_l}d(x_u,v_l)}\right\}.
\enas
The largest $\texttt{GD33}(k)$ indicates a valid optimal partition. 

\begin{flushleft}
\textbf{Point biserial correlation (PB) \cite{Mil80}}
\end{flushleft}
The point biserial correlation is a correlation-based index. A point biserial correlation is computed between corresponding entries in the original distance vector that contains distances between all pairs of points and the vector consisting of 0 and 1 entries that indicate whether two points are in the same cluster. The larger value of the point biserial correlation indicates a better partition.

\textbf{
\begin{flushleft}
PBM index \cite{PBM04}
\end{flushleft}
}

PBM index can be used with both crisp and fuzzy clustering algorithms. The crisp clustering version is defined as
\beas
\texttt{PBM}(k) = \left(\frac{1}{k} \times \frac{E_0}{E_k} \times D \right)^2 
\enas
\beas
\text{where \ }E_k = \sum_{i=1}^k \sum_{x \in C_i} d(x,v_i), \ \ E_0 = \sum_{j=1}^n d(x_j,\bar{x}) \text{ \ and \ } D = \max_{i,j=1}^k d(v_i,v_j).
\enas

It is based on the ratio of the total distance between each point and its centroid over the total distance between each point and the global centroid. The largest $\texttt{PBM}(k)$ indicates a valid optimal partition.

\begin{flushleft}
\textbf{Silhouette coefficient (SC) \cite{Rou87},\cite{KR90}}
\end{flushleft}
The silhouette value premiered in \cite{Rou87}. The term silhouette coefficient, which is the average silhouette values of all data points, was then used later in \cite{KR90}. The silhouette value measures the similarity of data points in the same cluster compared to other clusters. To define the silhouette, we begin with a few terms. For $x_i \in C_l$, let 
\beas
a(i) = \frac{1}{|C_l|-1}\sum_{\substack{j:x_j \in C_l \\ j\ne i} } d(x_i,x_j) \text{ \ and \ }
b(i) = \min_{s \ne l} \frac{1}{|C_s|}\sum_{j:x_j \in C_s} d(x_i,x_j).
\enas
Define the silhouette of a data point $x_i$ as
\beas
s(i) =  \begin{cases}
                    \frac{b(i)-a(i)}{\max\left\{a(i),b(i)\right\}}   \text{  \ \ if \ }  |C_l|>1  \\
                     0   \texttt{  \ \ \ \ \ \  \ \ \ \ \ \ \ \ \  \  if \  } |C_l|=1.
                 \end{cases} 
\enas

The silhouette coefficient is defined as 
\beas
\texttt{SC}(k) = \frac{1}{n} \sum_{i=1}^n s(i).
\enas
The largest $\texttt{SC}(k)$ indicates a valid optimal partition. 

\begin{flushleft}
\textbf{STR index \cite{Sta17}}
\end{flushleft}

The STR index was generalized from the PBM index in the crisp clustering-only version. It is defined as
\beas
STR(k) = \left[E(k)-E(k-1)\right]\left[D(k+1)-D(k)\right],
\enas
\beas
\text{where \ }D_{kmax} = \max_{i,j=1}^k d(v_i,v_j), \ \ D_{kmin} = \min_{i,j=1}^k d(v_i,v_j),
\enas
\beas
D(k) = \frac{D_{kmax}}{D_{kmin}} \text{ \ and \ } E(k) = \frac{E_0}{E_k}.
\enas
This index is based on the improvement from $k-1$ to $k$ clusters of the ratio of the total distance between each point and its centroid over the total distance between each point and the global centroid and the improvement from $k$ to $k+1$ of the ratio of the farthest and the closest centroids. The largest $\texttt{STR}(k)$ indicates a valid optimal partition.

\section{The proposed cluster validity indices}\label{sec:main}

As mentioned in the introduction, our index is motivated by the cophenetic correlation \cite{SR62} that computes the Pearson correlation between the vector of distances between all pairs of points and the vector containing the corresponding heights of the nodes at which the points are first joined together in the dendrogram. We replace the first joined height of a pair by the distance of the centroids of the corresponding clusters that the two points occupy.
We refer to our new correlation and new correlation index as NC and NCI, respectively. As mentioned in the introduction, our proposed index is intended for crisp clustering algorithms and requires only the data set and the clustering membership labels. Nevertheless, our index also works with fuzzy clustering algorithms when assigning data points to corresponding clusters using the maximum membership degree criterion.

For each data point $x_i$, $i=1,2,\ldots,n$, let $v_i(k)$ denote the centroid of the cluster that $x_i$ occupies from a clustering result with k groups and $v_0$ denote the centroid of the entire data. Let 
\bea \label{dmdef}
\vec{d}_v = (d(x_i,v_0))_{i \in [n]}
\ena
be a vector of length $n$ containing distances of all data points to the centroid of the entire data,
\bea \label{ddef}
\vec{d} = (d(x_i,x_j))_{i,j \in [n]}
\ena
be a vector of length ${n \choose 2}$ containing distances of all pairs of data points, and 
\bea \label{cdef}
\vec{c}(k) = (d(v_i(k),v_j(k)))_{i,j \in [n]}
\ena 
be a vector of the same length containing distances of all pairs of corresponding centroids of clusters that two points occupy. Though $d(\cdot,\cdot)$ can be any distances, we consider only the Euclidean distance in this entire work. Our correlation is defined as follows.

\begin{definition} \label{nc}
Given $\vec{d}$ and $\vec{c}(k)$ are as in \ref{ddef} and \ref{cdef}, repectively. NC correlation is defined as  
\beas
\texttt{NC}(k) = \Corr(\vec{d},\vec{c}(k)) 
\enas
for $k=2,3,\ldots,n$, and
\beas
\texttt{NC}(1) = 0  \text{ \ \ or \ \ } \texttt{NC}(1) = \frac{\texttt{SD}(\vec{d}_v)}{\max \vec{d}_v - \min \vec{d}_v}
\enas
where $\Corr(\cdot,\cdot)$ denotes the Pearson correlation, Kendall's Tau correlation, or Spearman's rank correlation. 
\end{definition}
We may write $\texttt{NC}_p$, $\texttt{NC}_k$, and $\texttt{NC}_s$ for Pearson, Kendall, and Spearman, respectively, but for simplicity, we will only use NC without subscripts to denote Pearson correlation throughout this paper. For Pearson correlation, 
\beas
\texttt{NC}(k) = \frac{\sum_{i,j\in [n]}(d_{ij}-\bar{d})((c_{ij}(k)-\bar{c}(k)))}{\sqrt{\sum_{i,j\in [n]}(d_{ij}-\bar{d})^2}\sqrt{\sum_{i,j\in [n]}(c_{ij}(k)-\bar{c}(k))^2}}.
\enas

We provide the user two options for $\texttt{NC}(1)$. If $\texttt{NC}(1)=0$ is chosen, then $\texttt{NC}(2)$ will improve noticeably from $\texttt{NC}(1)$, which will often result in our index providing a peak at $k=2$. Otherwise, $\texttt{NC}(1)$ is based on the ratio of the standard deviation and the difference between the maximum and minimum distance between each point and the centroid of the entire data set. Intuitively, if the SD is small compared to the difference between the maximum and minimum, dividing into two groups will yield better results.

Next, we state some properties of NC.

\begin{proposition}
For $k \in [n]$ and \texttt{NC}$(k)$ as in Definition \ref{nc},
\begin{enumerate}
	\item $-1\le \texttt{NC}(k) \le 1$
	\item \texttt{NC}$(n)=1$.
\end{enumerate}
\end{proposition}
\proof
The first item follows immediately from the fact that $\Corr$ is a correlation coefficient and that NC$(1) \ge 0$.
For the second item, when $k=n$ we have $\vec{c}(n)=\vec{d}$. Hence, NC$(n) = \Corr(\vec{d},\vec{d})=1$.
\bbox

A large NC (close to 1) indicates a good linear relationship between the true distance and the centroid cluster distance. Note that NC$(k) \rightarrow 1$ as $k \rightarrow n$. That is, if each point forms its own cluster, the linear association is perfect. Since this perfect association does not provide any information, we propose an adjustment to the NC correlation in Definition \ref{nci}.

For $k=2,3,\ldots,n-1$, we let
\bea \label{nci1}
\texttt{NCI1}(k) &=& \frac{\texttt{NC}(k)-\texttt{NC}(k-1)}{1-\texttt{NC}(k-1)}\Bigg/ \frac{\max\{0,\texttt{NC}(k+1)-\texttt{NC}(k)\}}{1-\texttt{NC}(k)} \nn \\
           &=& \frac{\left(\texttt{NC}(k)-\texttt{NC}(k-1)\right)\left(1-\texttt{NC}(k)\right)}{\max\{0,\texttt{NC}(k+1)-\texttt{NC}(k)\}\left(1-\texttt{NC}(k-1)\right)} 
\ena

and 

\bea \label{nci2}
\texttt{NCI2}(k) = \frac{\texttt{NC}(k)-\texttt{NC}(k-1)}{1-\texttt{NC}(k-1)} - \frac{\texttt{NC}(k+1)-\texttt{NC}(k)}{1-\texttt{NC}(k)}.
\ena

\begin{definition} \label{nci}
For $m=2,3,\ldots,n-1$, NC as in Definition \ref{nc} and NCI1 and NCI2 as in \eqref{nci1} and \eqref{nci2}, respectively, NCI is defined as

\begin{flushleft}
\textbf{Case 1}: $\max_{2\le l \le m} \texttt{NCI1}(k) < +\infty$
\end{flushleft}
\beas
\texttt{NCI}_m(k) =  
                 \begin{cases}
                     \min_{2\le l \le m} \left\{\texttt{NCI1}(l)|\texttt{NCI1}(l)> -\infty\right\}  \text{  \ \ if \ }  \texttt{NCI1}(k) = -\infty  \\
                     \texttt{NCI1}(k)   \text{ \ otherwise},
                 \end{cases} 
\enas
for $k=2,3,\ldots,m$.. 

\begin{flushleft}
\textbf{Case 2}: $\max_{2\le l \le m} \texttt{NCI1}(k) = +\infty$
\end{flushleft}
\beas
\texttt{NCI}_m(k) =  
                 \begin{cases}
                     
                     \min_{2\le l \le m} \left\{\texttt{NCI1}(l)|\texttt{NCI1}(l)> -\infty\right\} + \texttt{NCI2}(k)  \text{  \ \ if \ }  \texttt{NCI1}(k) = -\infty \\
										 \max_{2\le l \le m} \left\{\texttt{NCI1}(l)|\texttt{NCI1}(l)< +\infty\right\} + \texttt{NCI2}(k)  \text{  \ \ if \ }  \texttt{NCI1}(k) = +\infty \\
										 \texttt{NCI1}(k)+ \texttt{NCI2}(k)   \text{ \ otherwise},
                 \end{cases} 
\enas
for $k=2,3,\ldots,m$. 

\end{definition}

Now we give some insight onto NCI1 and NCI2. NCI1 and NCI2 are the proportion and the difference, respectively, of the same two ratios. The first ratio is the NC improvement from $k-1$ clusters to $k$ clusters over the entire room for improvement. The second ratio is the NC improvement from $k$ clusters to $k+1$ clusters over the entire room for improvement. We define the first ratio to see how much the proportion from $k$ clusters improve over $k-1$ compared to the entire room to improve. Having only the first ratio could lose some information. For instance, if increasing $k-1$ to $k$ clusters does not improve much compared to the entire room, but increasing $k$ to $k+1$ clusters even improves less,  $k$ clusters may be acceptable in this situation, but the first ratio will yield a small value at $k$. Dividing or subtracting by the second term overcomes this problem as it provides a local peak at $k$ in this instance. Therefore, if increasing $k-1$ to $k$ clusters does not improve much compared to the entire room, but it improves the most when compared to the numbers nearby, $k$ will be one of the candidates for the optimal number of clusters. 

The following remark explains the intuition behind our new cluster validity index given in Definition \ref{nci}.

\begin{remark}
\begin{enumerate}
	\item Notice that for \texttt{NCI1} we take $\max\{0,\texttt{NC}(k+1)-\texttt{NC}(k)\}$ to avoid the problem when the difference is negative, which can cause the proportion to be negative and large in magnitude when the ratio in the numerator is large. In this situation, $k$ should be a good candidate as $\texttt{NC}(k)$ improves a lot from $\texttt{NC}(k-1)$, and $\texttt{NC}(k+1)$ does not improve at all from $\texttt{NC}(k)$. 
	\item If there exists $l$ such that $\texttt{NCI1}(l) = +\infty$, we define $\texttt{NCI}_m(k)$ to be the sum of $\texttt{NCI2}(k)$ and the maximum of finite $\texttt{NCI1}(l)$ for $2 \le l \le m$. Intuitively, $k$ whose $\texttt{NCI1}(k) = +\infty$ are the final candidates, and we take the one with largest $\texttt{NCI2}(k)$.
	\item Our proposed index is based on \texttt{NCI1}. \texttt{NCI2} is only defined to handle the above situation.
	\item The highlighted benefit of our index is that it can provide many local peaks, which allow us to rank numbers of clusters that are potentially possible to choose. 
	\item Since NC is defined as the correlation between the true distance of a pair of data points and the distance between the centroids of clusters that this pair occupies, this correlation can also help evaluate whether the clustering result quality met a user's expectation. The value close to one indicates an excellent clustering quality. 
\end{enumerate}
\end{remark}

\section{Computational cost comparison} \label{sec:comp}

Table \ref{tabcomp} displays the computational cost comparison. Recall that $n$, $p$, and $k$ denote the data size a, the number of attributes, and the number of clusters, respectively. Note that the costs are computed under the Euclidean distance and Pearson correlation and the assumption that all $k$ clusters have size $n/k$, which yields the highest cost. 

\begin{table*}[h] 
\renewcommand{\arraystretch}{0.9}
\caption{Computational cost comparison}
\centering
\scriptsize	
\begin{tabular}{|c|c|c|c|} 
 \hline
 Index & multiplication & square root   \\ [0.5ex] 
 \hline
\texttt{NC}	& $\frac{(p+1)n(n-1)}{2}+\frac{pk(k-1)}{2}$ &	$\frac{n(n-1)}{2}$ \\ 
\texttt{NCI}	& $\frac{3(p+1)n(n-1)}{2}+\frac{3pk(k-1)}{2}+1$ &	$\frac{3n(n-1)}{2}$  \\ 
 \texttt{CH} &	$p(n+k)+2k+4$ &	$n+k$ \\
 \texttt{CSL} &	$\frac{pn}{2}\left(\frac{n}{k}-1\right)+\frac{pk}{2}(k-1)+k+1$ &	$\frac{n}{2}\left(\frac{n}{k}-1\right)+\frac{k}{2}(k-1)$ \\
 \texttt{DI} &	$\frac{pn}{2}\left(\frac{n}{k}-1\right)+\frac{p(k-1)n^2}{2k}+1$ &	$\frac{n}{2}\left(\frac{n}{k}-1\right)+\frac{(k-1)n^2}{2k}$ \\
 \texttt{DB} &	$pn+k+\frac{(p+1)k(k-1)}{2}+2$ &	$1$ \\
 \texttt{DB*} &	$pn+k+\frac{(p+1)k(k-1)}{2}+2$ &	$1$ \\
 \texttt{GD33} &	$\frac{pn^2}{k^2}+pn+2k+3$	& $\frac{n^2}{k^2}+n$  \\ 
 \texttt{PB} &	$\frac{(p+1)n(n-1)}{2}$ &	$\frac{n(n-1)}{2}$ \\ 
 \texttt{PBM} &	$2pn+\frac{pk(k-1)}{2}+k+5$ &	$2n+\frac{k}{2}(k-1)$ \\ 
 \texttt{SC} &	$\frac{pn^2(k-1)}{k}+pn\left(\frac{n}{k}-1\right)+k+2$ &	$\frac{n^2(k-1)}{k}+n\left(\frac{n}{k}-1\right)$  \\
 \texttt{STR} &	$4pn+pk(k-1)+2k+5$ &	$4n+k(k-1)$  \\[1ex] 
 \hline
\end{tabular}
\label{tabcomp}
\end{table*}

As we can see in the table, CH, DB, DB*, PBM, and STR are among the least expensive indices to compute with $O(np)$ multiplication operations and $O(n)$ square root operations. In contrast, NCI, NC, CSL, DI, GD33, PB, and SC require $O(n^2p)$ multiplication operations and $O(n^2)$ square root operations. Our index is slightly more expensive than most indices with the same order of operations since their costs have the factor of $3/2$ in the main terms. This cost is because our index is computed based on 3 NC correlations.

\section{Experimental Results}\label{sec:exp}

In this section, we compare the performance of our proposed validity index to the ten well-known ones detailed in Section \ref{sec:background}. Most of these indices were compared in \cite{CSL04}, \cite{AGMPP13}, and recently \cite{Sta17}. This paper tests these eleven indices in the following four subsections. The first subsection shows our highlighted advantage based on two artificial data sets. More artificial and real-world data sets appear in the second and third subsections, respectively. In the last subsection, we discuss an application in the business area. As for clustering tools, we apply either classic k-means clustering or hierarchical clustering (complete, average, or single linkage) with Euclidean distance, whatever is more appropriate for each data set. Note that all the validity indices are computed from the same clustering results. Though our proposed index applies to any crisp clustering algorithms, it is appropriate for data sets where each cluster is distributed around its mean, as our index is defined based on the correlation of distances. Hence it does not work well with some special types of data, such as data that includes spiral or ring clusters.

\subsection{Artificial Data Sets for showing the highlighted feature}

\begin{figure}[h]
    \centering
    \begin{subfigure}[t]{0.47\textwidth} 
		    \centering
        \includegraphics[width=1.8in]{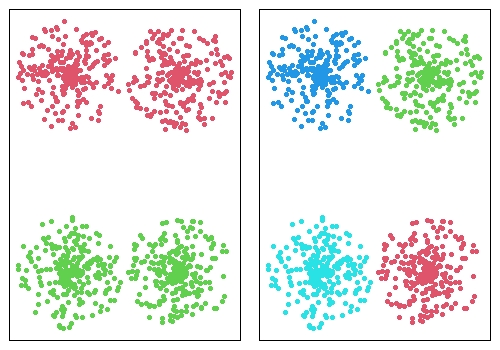}
        \caption{The data set from Fig.\ref{2or4}}
        \label{2or4-2}
    \end{subfigure}
    \begin{subfigure}[t]{0.47\textwidth} 
		     \centering
         \includegraphics[width=3in]{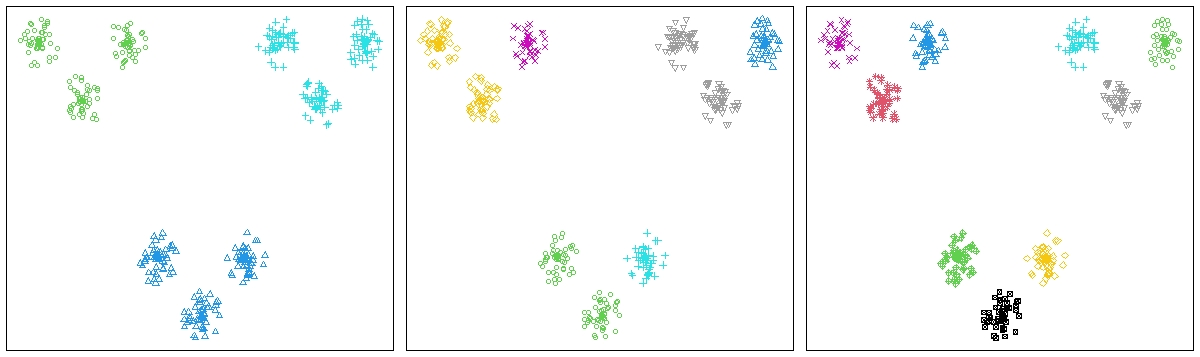}
         \caption{The data set from Fig.\ref{3or9}}
         \label{3or9-2}   
    \end{subfigure}
\caption{Examples to illustrate our proposed index's highlighted feature. The clustering results are obtained by k-means. }
\label{fig:main2}
\end{figure}

\begin{figure}[h]
    \centering
    \begin{subfigure}[t]{0.47\textwidth} 
        \includegraphics[width=2.5in]{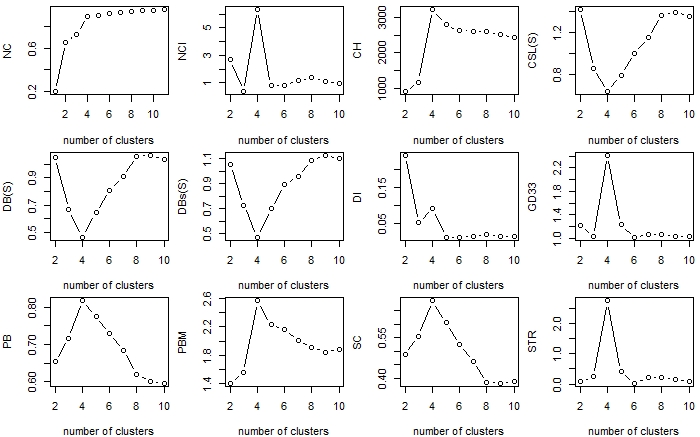}
        \caption{The data set from Fig.\ref{2or4}}
        \label{2or4-3}
    \end{subfigure}
    \begin{subfigure}[t]{0.47\textwidth} 
         \includegraphics[width=2.5in]{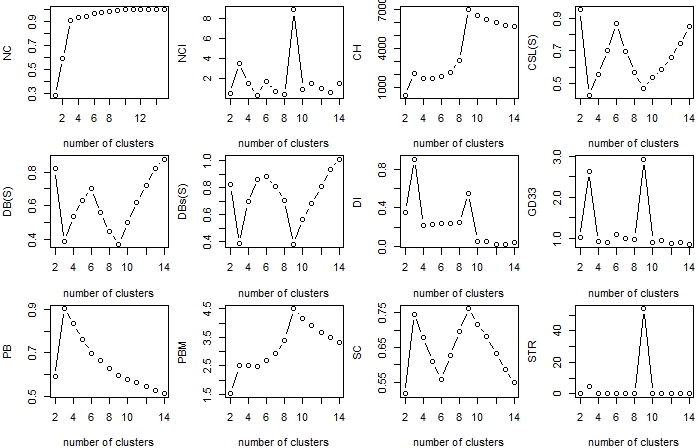}
         \caption{The data set from Fig.\ref{3or9}}
         \label{3or9-3}   
    \end{subfigure}
\caption{Cluster validity indices comparison. (S) is added on the y-axis to the indices where the smallest value indicates the optimum.}
\label{fig:cluscomp}
\end{figure}

In this subsection, we compare our index to the well-known indices to illustrate our highlighted feature, that it always provides sub-optimal numbers of clusters. The primary advantage is that it allows users to make decisions based on experiences and situations.

\begin{table*}[h] 
\renewcommand{\arraystretch}{0.5}
\caption{The optimal and sub-optimal numbers of clusters for data sets from Scenario 1 and 2 based on indices.}

\smallskip\noindent\setlength\tabcolsep{2pt} 
\begin{center}
\scalebox{0.7}{
\begin{tabularx}{\columnwidth}{@{}%
    L{1} C{1}C{1}C{1}C{1}C{1}@{}} 
\toprule
\mycell{CVI} & \multicolumn{2}{c@{}}{Scenario 1} & \multicolumn{3}{c@{}}{Scenario 2} \\
\cmidrule(l){2-3}
\cmidrule(l){4-6}
& 1st & 2nd  & 1st & 2nd & 3rd \\
\midrule 
    NCI        & 4 & 2 & 9 & 3 & 6 \\
    CH       & 4 & 5 & 9 & 10 & 11  \\
    CSL       & 4 & 5 & 3 & 9 & 10  \\
		DI       & 2 & 4 & 3 & 9 & 2  \\
		DB       & 4 & 5 & 9 & 3 & 8  \\
		DB*       & 4 & 5 & 9 & 3 & 10  \\
		GD33       & 4 & 5 & 9 & 3 & 6  \\
		PB       & 4 & 5 & 3 & 4 & 5  \\
		PBM       & 4 & 5 & 9 & 10 & 11  \\
		SC       & 4 & 5 & 9 & 3 & 10  \\
		STR       & 4 & 3 & 9 & 14 & 13  \\
\bottomrule
\label{sums1s2}
\end{tabularx}}
\end{center}
\end{table*}

The first scenario is from Example \ref{24ex} in Section \ref{sec:introduction}. As mentioned there, partitioning into four clusters is preferable, but two clusters are also a good option, as shown in Fig. \ref{2or4-2}. Obtaining three clusters by combining either two spheres on top or below is acceptable, but if we combine two clusters on top, it seems unreasonable not to combine the bottom one too. As shown in Fig. \ref{2or4-3} and Table \ref{sums1s2}, all indices except DI indicate that four is the optimal number of clusters. While other indices provide only one peak at four, our NCI also provides a sub-optimal option at two with $\texttt{NCI}_{10}(2)=2.688$ and $\texttt{NCI}_{10}(4)=6.325$. Moreover, the NC correlation also allows us to check whether two or four clustering quality meets our expectation with NC$(2)=0.654$ and NC$(4)=0.890$.

The second scenario is from Example \ref{39ex} in Section \ref{sec:introduction}. As mentioned there, partitioning into nine clusters is preferable, but three and six clusters are second and third options, respectively. See Fig. \ref{3or9-2} for the clustering results with $k=3, 6$, and $9$. As shown in Fig. \ref{3or9-3} and Table \ref{sums1s2}, CH and PBM provide only one peak at nine, while DB, DB*, GD33, SC, STR, and our NCI indicate nine as the optimal number of clusters and three as the second optimum. Conversely,  CSL, DI, and PB indicate that three is the optimal number of clusters, and nine is the second optimum. However, NCI is the only one that also gives a local peak at six clusters. The values of our index are $\texttt{NCI}_{14}(3)=3.573$, $\texttt{NCI}_{14}(6)=1.765$, and $\texttt{NCI}_{14}(9)=8.856$. 

The two scenarios emphasize that our NCI provides more than one option when appropriate as claimed.

\subsection{More Artificial Data Sets}

\begin{table*}[h] 
\renewcommand{\arraystretch}{0.5}
\caption{Artificial data sets description. The data sets are available in \cite{Benchmark} except for D1-D3.}

\smallskip\noindent\setlength\tabcolsep{1pt} 
\begin{center}
\scalebox{0.7}{
\begin{tabularx}{\columnwidth}{@{}%
    L{1} C{1}C{1}C{1}C{1}@{}} 
\toprule
\mycell{Data} & Source  & {\# instance} & {\# features} & {\# classes} \\
\midrule
     D1  &\cite{CSL04}        & 650  & 2 & 3   \\
		 D2 &\cite{CSL04}         & 620  & 2 & 3  \\
		 D3 &\cite{CSL04}         & 500  & 2 & 5     \\
		 D4 &\cite{D4},\cite{D4-2}     & 900 & 2 & 9  \\
	 	 D5 &\cite{D5}        & 3000  & 2 & 9   \\
		 D6 &\cite{D6}        & 600  & 2 & 15  \\
		 D7 &\cite{D7}       & 1000  & 2 & 2  \\
		 D8 &\cite{D7}       & 1000  & 2 & 2  \\
		 D9 &\cite{D7}        & 1000  & 2 & 4   \\
		 D10 &\cite{D10}        & 863  & 2 & 4  \\
		 D11 &\cite{D10}        & 1261  & 2 & 4  \\
		 D12 &\cite{Benchmark}      & 2000  & 2 & 3  \\
		 D13 &\cite{D13},\cite{D13-2},\cite{D4-2}        & 500  & 2 & 10   \\
		 D14 &\cite{D14}       & 512  & 2 & 4 \\
\bottomrule
\label{dataarf}
\end{tabularx}}
\end{center}
\end{table*}

\begin{figure}[H]
    \centering
    \begin{subfigure}[t]{0.47\textwidth} 
        \includegraphics[width=2.5in]{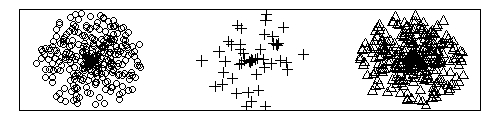}
        \caption{}
        \label{ex1}
    \end{subfigure}
    \begin{subfigure}[t]{0.47\textwidth} 
         \includegraphics[width=2.5in]{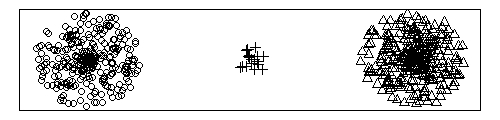}
         \caption{}
         \label{ex2}   
    \end{subfigure}
		\begin{subfigure}[t]{0.47\textwidth} 
         \includegraphics[width=2.5in]{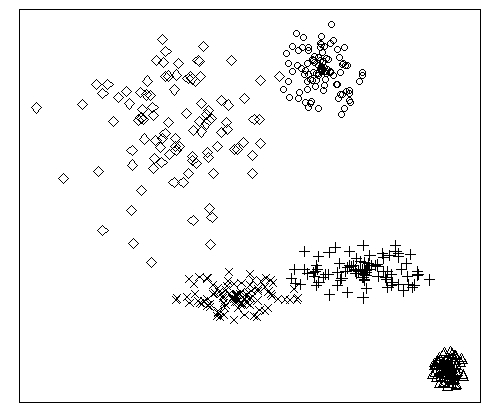}
         \caption{}
         \label{ex3}   
    \end{subfigure}
		\begin{subfigure}[t]{0.47\textwidth} 
         \includegraphics[width=2.5in]{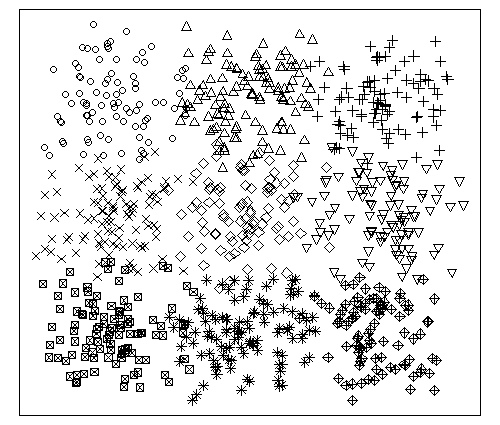}
         \caption{}
         \label{st900}   
    \end{subfigure}
		\begin{subfigure}[t]{0.47\textwidth} 
         \includegraphics[width=2.5in]{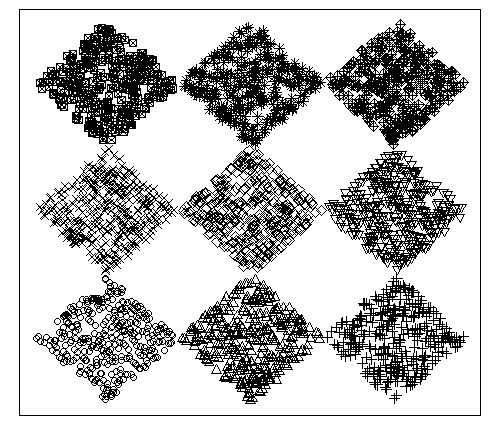}
         \caption{}
         \label{diamond9}   
    \end{subfigure}
		\begin{subfigure}[t]{0.47\textwidth} 
         \includegraphics[width=2.5in]{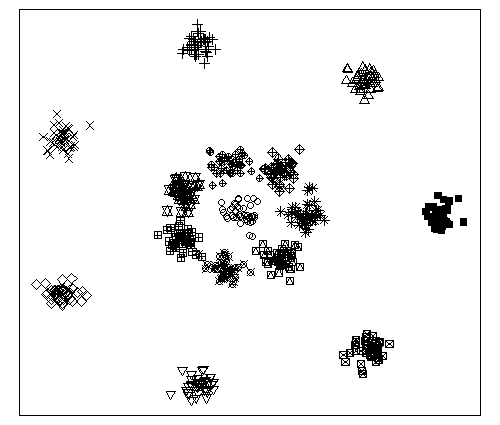}
         \caption{}
         \label{R15}   
    \end{subfigure}
\caption{More artificial examples}
\label{fig:main3}
\end{figure}

\begin{figure}[H]
    \centering
   	\begin{subfigure}[t]{0.47\textwidth} 
         \includegraphics[width=2.5in]{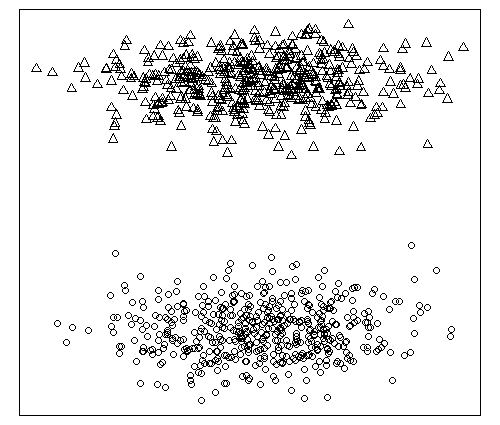}
         \caption{}
         \label{long1}   
    \end{subfigure}
		\begin{subfigure}[t]{0.47\textwidth} 
         \includegraphics[width=2.5in]{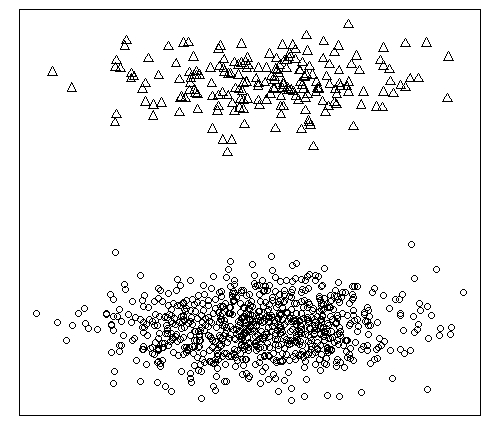}
         \caption{}
         \label{long3}   
    \end{subfigure}
		\begin{subfigure}[t]{0.47\textwidth} 
         \includegraphics[width=2.5in]{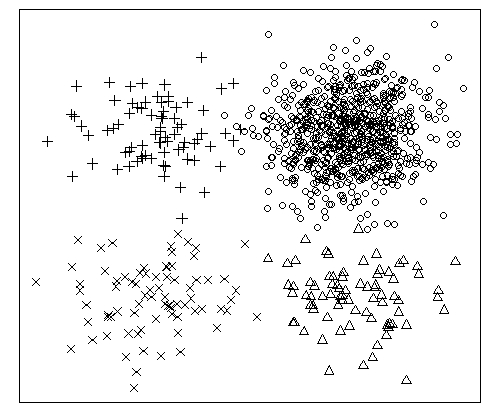}
         \caption{}
         \label{sizes5}   
    \end{subfigure}
		\begin{subfigure}[t]{0.47\textwidth} 
         \includegraphics[width=2.5in]{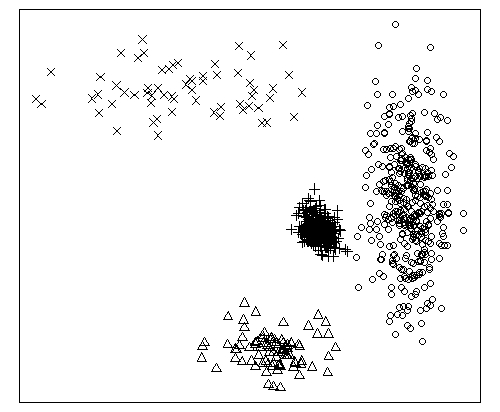}
         \caption{}
         \label{2d-4c-no4}   
    \end{subfigure}
		\begin{subfigure}[t]{0.47\textwidth} 
         \includegraphics[width=2.5in]{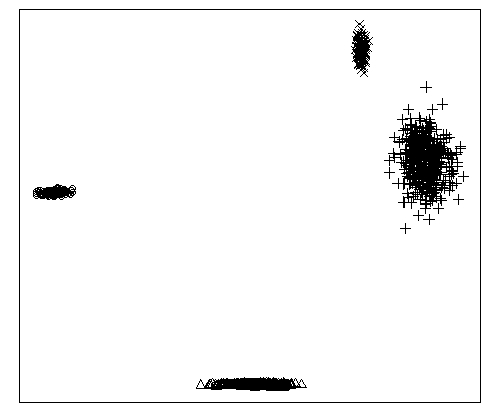}
         \caption{}
         \label{2d-4c}   
    \end{subfigure}
		\begin{subfigure}[t]{0.47\textwidth} 
         \includegraphics[width=2.5in]{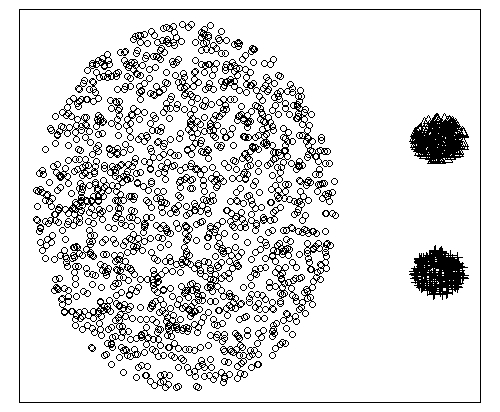}
         \caption{}
         \label{cure-t0-2000n-2D}   
    \end{subfigure}
\caption{More artificial examples}
\label{fig:main4}
\end{figure}

\begin{figure}[H]
    \centering
		\begin{subfigure}[t]{0.47\textwidth} 
         \includegraphics[width=2.5in]{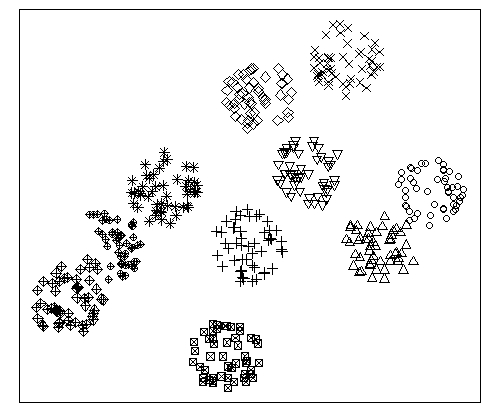}
         \caption{}
         \label{elliptical_10_2}   
    \end{subfigure}
		\begin{subfigure}[t]{0.47\textwidth} 
         \includegraphics[width=2.5in]{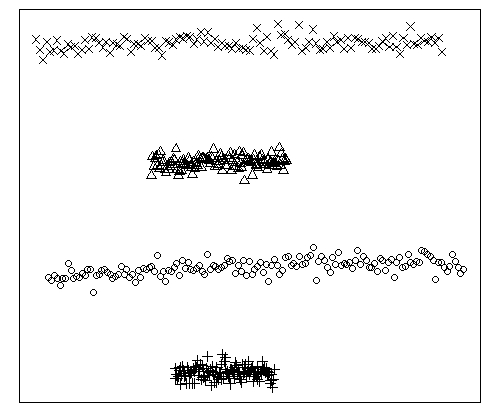}
         \caption{}
         \label{zelnik}   
    \end{subfigure}
\caption{More artificial examples}
\label{fig:main5}
\end{figure}

In this subsection, we compare our index to the other well-known indices using 14 benchmark artificial data sets to illustrate that our index can accurately handle clusters with different sizes, densities, and shapes. The data sets are detailed in Table \ref{dataarf} and plotted in Figures \ref{fig:main3}, \ref{fig:main4}, and \ref{fig:main5}. We intentionally choose the data sets which do not contain either rings or spirals, as mentioned earlier.

Based on our intention to introduce a new index that can provide not only the optimal but also sub-optimal numbers of clusters. The appropriate judgment of the indices should be the rank of the actual number of classes shown in Table \ref{summaryarf3}. That is, if the index could not detect the number of classes as the optimum, it should be at least highly ranked. For instance, it will show 1 if the index can detect the true number of classes, show 2 if the index has the actual number of classes as the second optimum and show 5, if the actual number of classes is ranked fifth based on the index. The optimal number of clusters for each data set according to each cluster validity index appears in Table \ref{summaryarf2}. 

\begin{table*}[h] 
\renewcommand{\arraystretch}{0.5}
\caption{The rank of the optimal number of clusters for each artificial data set based on each cluster validity index. The Mtd column is the clustering method: k=k-means and hs=single linkage hierarchical clustering. The Scl column is the normalizing method: sc=standard score and no=no normalizing. The m column is the maximum number of clusters considered. The row $\bar{x}$ shows the average rank of the actual number of classes.}

\smallskip\noindent\setlength\tabcolsep{1pt} 
\begin{center}
\scalebox{0.7}{
\begin{tabularx}{\columnwidth}{@{}%
    L{1} C{1}C{1}C{1}C{1}C{1}C{1}C{1}C{1}C{1.2}C{1}C{1}C{1}C{1}C{0.8}@{}} 
\toprule
\mycell{CVI}  & {NCI} & {CH} & {CSL} & {DB} & {DB*} & {D} & {GD} & {PB} & {PBM} & {SC} & {STR} & {Mtd} & {Scl} & {m}  \\
\midrule
     D1       & 1 & 1 & 1 & 2 & 2 & 1 & 2 & 2  & 1 & 2 & 1 & k & no & 10   \\
     D2       & 1 & 2 & 1 & 1 & 2 & 1 & 2 & 2  & 2 & 2 & 1  & k & no & 10  \\
     D3       & 1 & 4 & 1 & 1 & 1 & 8 & 4 & 3  & 1 & 1 & 1 & k & no & 10  \\
		 D4       & 1 & 1 & 1 & 1 & 1 & 1 & 1 & 7  & 1 & 1 & 1 & k & no & 12    \\
		 D5       & 1 & 1 & 1 & 1 & 1 & 1 & 1 & 7  & 1 & 1 & 1 & k & no & 12  \\
		 D6       & 1 & 1 & 4 & 1 & 1 & 5 & 1 & 12 & 1 & 1 & 1 & k & no & 20 \\
		 D7       & 1 & 9 & 1 & 1 & 1 & 1 & 2 & 3  & 8 & 2 & 7 & k & sc & 10 \\
		 D8       & 1 & 9 & 9 & 9 & 2 & 1 & 1 & 1  & 3 & 1 & 3 & k & sc & 10 \\
		 D9       & 5 & 7 & 1 & 1 & 1 & 2 & 1 & 1  & 3 & 1 & 1 & k & no & 10   \\
		 D10      & 1 & 7 & 6 & 1 & 1 & 3 & 3 & 1  & 3 & 1 & 1 & k & no & 10  \\
		 D11      & 1 & 7 & 1 & 1 & 1 & 3 & 2 & 2  & 1 & 1 & 1 & k & no & 10  \\
		 D12      & 1 & 1 & 4 & 1 & 2 & 2 & 2 & 7  & 3 & 1 & 1 & k & sc & 10   \\
		 D13      & 1 & 1 & 2 & 3 & 1 & 5 & 2 & 9  & 1 & 1 & 1 & k & no & 15  \\
		 D14      & 1 & 3 & 9 & 6 & 6 & 3 & 3 & 6  & 2 & 2 & 1 & hs & no & 10 \\
		 
		 $\bar{x}$  & \textbf{\underline{1.29}}  & 3.86  & 3.00 & 2.14 & 1.64  & 2.64 & 1.93  & 4.5 & 3.00  & \textbf{\underline{1.29}}  & \textbf{1.57 }   \\
\bottomrule
\label{summaryarf3}
\end{tabularx}}
\end{center}
\end{table*}

\begin{table*}[h] 
\renewcommand{\arraystretch}{0.5}
\caption{The optimal number of clusters for each artificial set based on each index. The last row shows the proportion of correct class detection.}

\smallskip\noindent\setlength\tabcolsep{2pt} 
\begin{center}
\scalebox{0.7}{
\begin{tabularx}{\columnwidth}{@{}%
    L{0.8} C{1}C{1}C{1}C{1}C{1}C{1}C{1}C{1}C{1}C{1.2}C{1}C{1}@{}} 
\toprule
\mycell{CVI} & {\#C} & {NCI} & {CH} & {CSL} & {DB} & {DB*} & {D} & {GD} & {PB} & {PBM} & {SC} & {STR}  \\
\midrule
     D1       & 3  & 3  & 3  & 3  & 2 & 2  & 3 & 2  & 2 & 3  & 2  & 3   \\
     D2       & 3  & 3  & 2  & 3  & 3 & 2  & 3 & 2  & 2 & 2  & 2  & 3   \\
     D3       & 5  & 5  & 8  & 5  & 5 & 5  & 2 & 4  & 4 & 5  & 5  & 5   \\
		 D4       & 9  & 9  & 9  & 9  & 9 & 9  & 9 & 9  & 3 & 9  & 9  & 9   \\
		 D5       & 9  & 9  & 9  & 9  & 9 & 9  & 9 & 9  & 4 & 9  & 9  & 9   \\
		 D6       & 15 & 15 & 15 & 8  & 8 & 15 & 8 & 15 & 8 & 15 & 15 & 15  \\
		 D7       & 2  & 2  & 10 & 2  & 2 & 2  & 2 & 4  & 4 & 7  & 4  & 6   \\
		 D8       & 2  & 2  & 10 & 9  & 3 & 4  & 2 & 2  & 2 & 5  & 2  & 5   \\
		 D9       & 4  & 10  & 6  & 4  & 4 & 4  & 3  & 4 & 4 & 5  & 4  & 4   \\
		 D10      & 4  & 4  & 7  & 6  & 4 & 3  & 2 & 2  & 4 & 6  & 4  & 4   \\
		 D11      & 4  & 4  & 9  & 4  & 4 & 4  & 3 & 3  & 3 & 4  & 4  & 4   \\
		 D12      & 3  & 3  & 3  & 5  & 3 & 2  & 2 & 2  & 10 & 2  & 3  & 3   \\
		 D13      & 10 & 10 & 10 & 9  & 9 & 10 & 11& 2  & 2 & 10 & 10 & 10   \\
		 D14      & 4  & 4  & 2  & 10 & 2 & 2  & 2 & 2  & 2 & 3  & 2  & 4    \\
		 Prop       &    & \textbf{\underline{0.93}}  & 0.43  & 0.57 & 0.71 & 0.64  & 0.43 & 0.36  & 0.21 & 0.50  & 0.71  & \textbf{0.86}   \\
\bottomrule
\label{summaryarf2}
\end{tabularx}}
\end{center}
\end{table*}

We show the average rank of the actual number of classes and the proportion of correct class detection for each index in Tables \ref{summaryarf3} and \ref{summaryarf2}, respectively. Our index performs best in both the proportion (0.93) and the average rank (1.29). STR index is the second in terms of proportion (0.86), and the silhouette coefficient also has the best average rank at 1.29.

\subsection{Real-World Data Sets}

\begin{table*}[h] 
\renewcommand{\arraystretch}{0.5}
\caption{UCI Real-world data sets description}

\smallskip\noindent\setlength\tabcolsep{1pt} 
\begin{center}
\scalebox{0.7}{
\begin{tabularx}{\columnwidth}{@{}%
    L{1.6} C{0.8}C{0.8}C{0.8}@{}} 
\toprule
\mycell{Data}  & {\# instance} & {\# features} & {\# classes} \\
\midrule
     Breast Cancer WI(BcW)          & 683  & 9 & 2   \\
		 Dermatology(Dem)          & 366  & 35 & 6  \\
		 Glass(Gls)          & 214  & 9 & 6     \\
		 Haberman(Hab)          & 306  & 3 & 2  \\
	 	 Ionosphere(Ion)          & 351  & 35 & 2   \\
		 Iris         & 150  & 4 & 3  \\
		 Parkinsons(Pak)           & 195  & 22 & 2  \\
		 Pendigits(PD)           & 10992  & 16 & 10  \\
		 Vertebral Column(VC)           & 310  & 6 & 3   \\
		 Wine         & 178  & 13 & 3  \\

\bottomrule
\label{datarw}
\end{tabularx}}
\end{center}
\end{table*}

In this subsection, we test our proposed index based on ten real-world UCI data sets \cite{UCI} detailed in Table \ref{datarw}.

\begin{table*}[h] 
\renewcommand{\arraystretch}{0.5}
\caption{The rank of the optimal number of clusters for each real-world data set based on each cluster validity index.  The Mtd column is the clustering method: k=k-means and hc=complete linkage hierarchical clustering. The Scl column is the normalizing method: sc=standard score and mm=min-max scaling. The m column is the maximum number of clusters considered. The row $\bar{x}$ shows the average rank of the actual number of classes. The last row shows the proportion of correct class detection.}

\smallskip\noindent\setlength\tabcolsep{1pt} 
\begin{center}
\scalebox{0.7}{
\begin{tabularx}{\columnwidth}{@{}%
    L{1} C{1}C{1}C{1}C{1}C{1}C{1}C{1}C{1}C{1.2}C{1}C{1}C{1}C{1}C{0.8}@{}} 
\toprule
\mycell{CVI}  & {NCI} & {CH} & {CSL} & {DB} & {DB*} & {D} & {GD} & {PB} & {PBM} & {SC} & {STR} & {Mtd} & {Scl} & {m}  \\
\midrule
     BcW          & 2  & 1 & 1  & 1  & 1  & 3 & 1   & 3  & 1  & 1  & 1 & k & mm & 6\\
		 Dem          & 2  & 5 & 4  & 5  & 5  & 3 & 4   & 3  & 5  & 5  & 2  & k & sc & 10 \\
		 Gls          & 3  & 7 & 9  & 1  & 1  & 3 & 1   & 2  & 3  & 7  & 4 & hc & sc & 10   \\
		 Hab          & 1  & 5 & 1  & 1  & 1  & 1 & 1   & 2  & 1  & 1  & 3  & k & mm & 6 \\
	 	 Ion          & 1  & 1 & 4  & 3  & 1  & 2 & 2   & 2  & 1  & 2  & 1  & k & sc & 6 \\
		 Iris         & 1  & 2 & 2  & 2  & 2  & 6 & 2   & 2  & 2  & 2  & 3 & k & sc & 7\\
		 Pak          & 4  & 1 & 5  & 5  & 1  & 5 & 1   & 5  & 5  & 1  & 1 & k & sc & 6 \\
		 PD           & 2  & 9 & 5  & 6  & 4  & 4 & 9   & 7  & 9  & 4  & 4  & k & sc & 14 \\
		 VC           & 3  & 2 & 5  & 4  & 4  & 6 & 5   & 1  & 3  & 2  & 4  & k & mm & 7 \\
		 Wine         & 1  & 1 & 3  & 1  & 1  & 1 & 2   & 2  & 1  & 1  & 1 & k & sc & 7 \\
	   $\bar{x}$  & \textbf{\underline{2.00}}	& 3.40	& 3.90	& 2.90	& \textbf{2.10} &	3.40 &	2.80 &	2.90 &	3.10 &	2.60 &	2.40   \\
		 Prop         & \textbf{0.4}  & \textbf{0.4} & 0.2  & \textbf{0.4}  & \textbf{\underline{0.6}}  & 0.2 & \textbf{0.4}   & 0.1  & \textbf{0.4}  & \textbf{0.4}  & \textbf{0.4 }\\
\bottomrule
\label{summaryrw}
\end{tabularx}}
\end{center}
\end{table*}

Since it is much more challenging to detect the true number of classes of the real-world data sets, we show only the rank of the actual number of classes based on each index in Table \ref{summaryrw}. For each data set, we define the maximum number of clusters to consider as four plus the true number of classes. As shown in Table \ref{summaryrw}, our index performs best based on the average rank (2.00), and Davies-Bouldin* index is second (2.10). However, the Davies-Bouldin* index gives the best proportion of 0.6 of true detection compared to our index of 0.4.

\subsection{Online Retail Data Set}

\begin{figure}[h]
\centering
\includegraphics[width=3in]{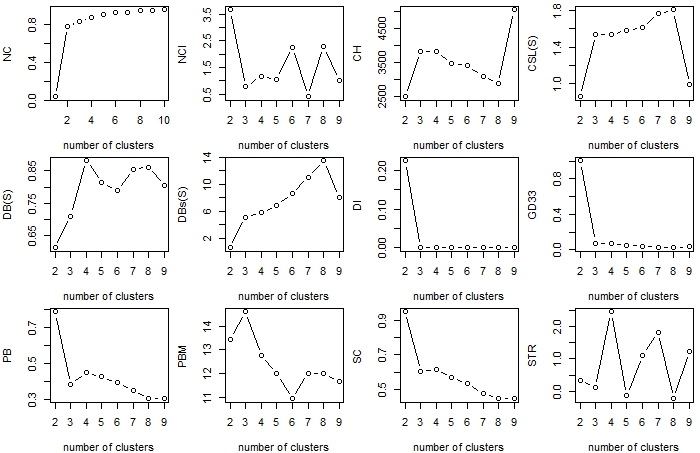}
\caption{The cluster validity indices comparison of the Online Retail Data Set from UCI achieved by k-means with $k=2$ to $9$.}
\label{mkt-3}
\end{figure}

In this subsection, we consider a data set modified from the UCI online retail data set \cite{UCI} consisting of 4,472 rows and three columns, where the features are RFM (Recency, Frequency, and Monetary.) Recall that the RFM method is well-known in marketing, where recency indicates the time since the last purchase, frequency counts the number of transactions, and monetary value is the total amount spent. Here the true number of classes is unknown, and an experienced analyst should decide the number of clusters. We use k-means with $k= 2, 3, \ldots, 9$ and apply
the eleven validity indices. The numerical values of all indices appear in Figure \ref{mkt-3}.

\begin{figure}[h]
\centering
\includegraphics[width=4.5in]{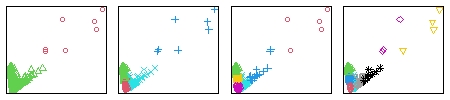}
\caption{The clustering results achieved by k-means with $k=2,4,6,8$ of the Online Retail Data Set from UCI where the x-axis and y-axis are the first and second principal components of the data.}
\label{mkt-2}
\end{figure}

We can see from Figure \ref{mkt-3} that most indices including ours indicate two as the optimal number of clusters. However, our index and STR are the only ones that provide several sub-optimal options. Our index gives four peaks at two, eight, six, and four, respectively. The clustering results for $k=2,4,6,8$ are shown in Figure \ref{mkt-2}. The x-axis and y-axis are the first principal component (mostly representing frequency and monetary value) and the second principal component (mostly representing recency), respectively. 

For two clusters, the first group is the group that makes the purchase more recently, more frequently, and with higher monetary values. For the four, six, and eight cluster results, we can see from Fig. \ref{mkt-2} that the groups that line up from top to bottom next to the y-axis have almost the same first principal component, but they are split into levels based on the second principal component. These groups can be characterized as regular customers (average spending amount and frequency) with different recent times of making a purchase. As for the remaining groups, they are split based on the first component. These groups spend more consistently with higher volume and thus could be considered loyal to this online retailer. They can be above average, high, and extremely high spending customers, for instance. 

To apply cluster analysis in marketing, it is always better to have several options for the number of clusters to choose from, as the true number of classes does not exist and is open to an experienced analyst. Therefore, our index is one of the best in this area.

\section{Conclusion} \label{sec:conclude}
In this work, we introduce a cluster validity index that can handle clusters with different sizes, densities, and shapes and always provides more than one choice of an optimal number of clusters. To be more precise, our index indicates not only the optimal number of clusters but also provides the sub-optimal numbers of clusters. This feature is very powerful in current applications, as in some areas, there is no absolute correct number of clusters. Another advantage of the proposed index is that they are defined very simply and easy to use, yet its effectiveness outperforms the well-known validity indices in many perspectives. An R package called NCvalid, which has several functions that compute our proposed index and other well-known validity indices mentioned in this work, is available at \url{https://github.com/nwiroonsri/NCvalid}.

Our proposed index has two drawbacks. First, it has high computational costs when $n$ is large as it must compute the distances between all ${n \choose 2}$ pairs of data points and all pairs of clusters and then the correlation between them. However, their computational costs are still similar to some known indices with a slightly larger constant. Secondly, our index only works with traditional crisp clustering methods. Hence, it only works with data sets compatible with those methods. Our index works with fuzzy clustering only when we assign data points to corresponding clusters using the maximum membership degree criterion. 

Now there are several deep learning-based clustering methods. One of the approaches is to perform traditional methods such as k-means after the original data points have been learned and transformed into a good feature representation (see\cite{YFSH17}.)  Another emerging state-of-the-art method is unsupervised 3D point cloud segmentation (see \cite{XTZ20} for a review), where traditional clustering algorithms such as k-means and fuzzy k-means are used in the early stage of some approaches. Following these two concepts, our cluster validity index can also detect the number of clusters at the stage that uses the traditional clustering algorithm.   

Future work will focus on generalizing our index to be compatible with fuzzy clustering methods and using our index in the traditional clustering stage of deep learning-based clustering methods. 

\section*{Acknowledgment}

The author would like to thank anonymous reviewers, the editor-in-chief and the associate editor for extremely helpful comments and detailed suggestions that led to a big improvement of the paper.





\end{document}